\documentclass[margin=3in]{article}

\usepackage{PRIMEarxiv}
\usepackage{comment}
\usepackage{hyperref}      
\usepackage{nicefrac}       
\usepackage{fancyhdr}       
\usepackage{graphicx}       
\usepackage{caption} 
\usepackage{algorithm}
\usepackage{algorithmic}

\usepackage{amsfonts}
\usepackage{subcaption}
\usepackage{multirow}
\usepackage{bm}
\usepackage[utf8]{inputenc}
\usepackage[autostyle]{csquotes}
\usepackage{comment}
\usepackage{mathtools}
\usepackage{xfrac}
\usepackage{bm}
\usepackage{etoolbox}
\usepackage{booktabs,array}

\usepackage{soul}
\usepackage{url}
\usepackage{xcolor}

\usepackage{graphicx}
\usepackage{amsmath}
\usepackage{amsthm}
\usepackage{booktabs}
\usepackage{algorithm}
\usepackage{algorithmic}
\usepackage{amsfonts}
\usepackage{subcaption}
\usepackage{multirow}
\usepackage{bm}
\usepackage[autostyle]{csquotes}
\usepackage{comment}
\usepackage{mathtools}
\usepackage{bm}
\usepackage{etoolbox}
\usepackage{booktabs,array}
\usepackage{amssymb}

\def \bim3d{\texorpdfstring{$BIM^{Sim}_{3D}$}~}
\newcommand{\simsis}{$\text{SIM}_{\text{sis}}$}
\newcommand{\ie}{i.e.,~}
\newcommand{\eg}{e.g.,~}

\DeclareMathOperator*{\argmax}{arg\,max}

\DeclarePairedDelimiter\ceil{\lceil}{\rceil}

\title{Grey-box Bayesian Optimization for Sensor Placement in Assisted Living Environments
}

\author{
  Shadan Golestan, Omid Ardakanian, Pierre Boulanger \\
  Department of Computing Science \\
  University of Alberta \\
  \texttt{\{golestan, ardakanian, pierreb\}@ualberta.ca} \\
}

\begin{document}
\maketitle

\begin{abstract}
Optimizing the configuration and placement of sensors is crucial for reliable fall detection, indoor localization, and activity recognition in assisted living spaces.
We propose a novel, sample-efficient approach to find a high-quality sensor placement
in an arbitrary indoor space based on grey-box Bayesian optimization and simulation-based evaluation.
Our key technical contribution lies in capturing domain-specific knowledge 
about the spatial distribution of activities
and incorporating it into the iterative selection of query points in Bayesian optimization.
Considering two simulated indoor environments and a real-world dataset 
containing human activities and sensor triggers,
we show that our proposed method performs better compared to state-of-the-art black-box optimization techniques in identifying high-quality
sensor placements, leading to accurate activity recognition in terms of F1-score, 
while also requiring a significantly lower (51.3\% on average) number of expensive function queries.
\end{abstract}


\section{Introduction}
Smart indoor spaces are commonly equipped with various networked sensors, such as
motion sensors and cameras, to monitor the activities and physical and mental health of the occupants~\cite{cook2012casas}.
Previous work shows that the placement of these sensors in the environment is one of the main factors
determining the performance of machine learning models that utilize the data generated by these sensors~\cite{pereira2018influence,yang2021adaptive}.
In particular, an optimized sensor placement strategy 
makes possible accurate activity recognition and localization using the smallest number of sensors.
However, due to the exponential size of the search space and the high cost of evaluating
the model performance for each sensor placement under normal occupancy conditions, 
finding the best sensor placement through exhaustive search is prohibitive in practice.

To date, many efforts have been made to design sample-efficient techniques to optimize the location of sensors, 
given a lower bound on the performance of downstream applications that consume the sensor data.
Greedy and evolutionary algorithms, such as the genetic algorithm~(GA), 
are the most notable methods used to place and configure sensors in an indoor environment~\cite{thomas2016genetic,wu2020sensor,yu2020optimizing}.
But these methods do not perform well in general, 
because they rely only on local information provided by the samples of $f$, 
the function being optimized~\cite{mori2005comparison},
with $x$ being a sensor placement and $f(x)$ representing some performance 
measure of the machine learning model.
On the contrary, estimation of distribution algorithms, such as Bayesian Optimization~(BO)~\cite{shahriari2015taking}, 
use local information to acquire global information about $f(x)$, 
\ie building a probabilistic surrogate model of this function. 
When this model represents $f(x)$ accurately, 
the optimizer can perform more effective exploration/exploitation.
A wide range of problems have been recently solved using BO, 
from
optimization over permutation spaces~\cite{deshwal2022bayesian}
and combinatorial spaces~\cite{deshwal2020optimizing,deshwal2023bayesian} 
to setting up sensor networks for air quality monitoring~\cite{hellan2022bayesian}.
However, the application of BO to optimize the sensor placement 
for indoor activity recognition has not been explored in previous work.

The main shortcoming of BO is that it treats $f(x)$ as a black-box function, 
meaning that it analyzes only its input-output behavior,
disregarding any inherent, domain-specific knowledge that might exist about this function. 
Grey-box optimization~\cite{astudillo2021thinking}, however,
incorporates this knowledge in the optimization process.
Although it requires more careful design, 
it can be faster and drastically improve the quality of the solution.
Our hypothesis is that in the problem of optimizing sensor placement
with respect to the performance of an activity recognition model, 
inherent knowledge about the distribution of activities in different parts of the space
could help BO quickly identify important regions in the search space. 

\begin{figure}[tb]
    \centering
    \centerline{\includegraphics[width=0.6  \columnwidth]{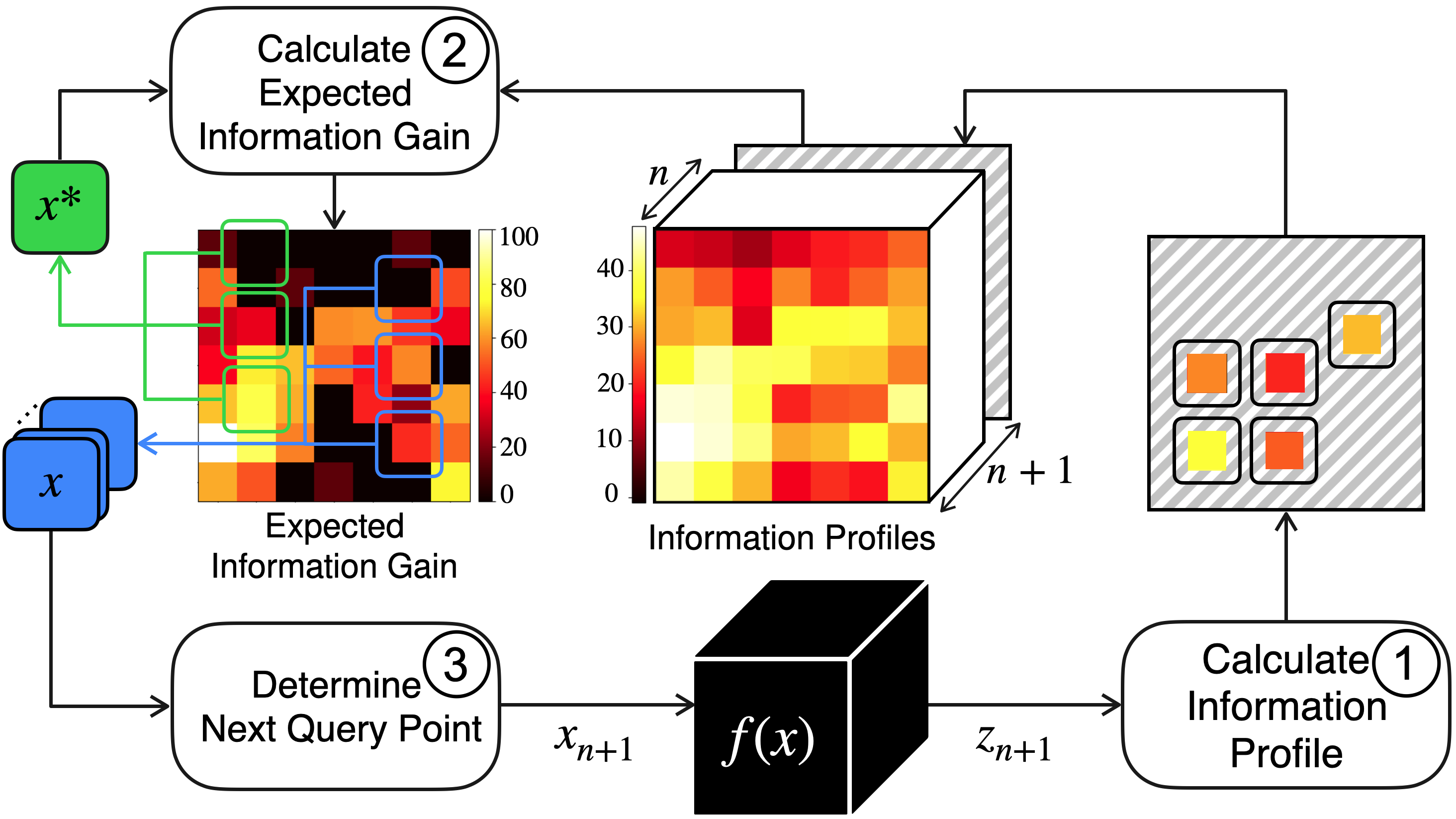}}
    \caption{An overview of our proposed grey-box optimization algorithm (DGBO).}
    \label{intro_fig}
\end{figure}

This paper presents Distribution-Guided Bayesian Optimization (DGBO),
a novel grey-box BO algorithm that learns the spatial distribution of activities 
and takes advantage of this knowledge to speed up the search for the best sensor placement.
This algorithm, depicted in Figure~\ref{intro_fig},
combined with high-fidelity building simulation for 
generating realistic movement and activity traces and corresponding sensor triggers,
establishes the foundation of a framework for sample-efficient identification 
of a motion sensor placement that supports accurate activity recognition in an aged-care facility.
The proposed approach does not cause any discomfort for the occupants,
nor does it infringe their privacy or raise major ethical concerns.
We show empirically that DGBO outperforms BO, genetic, and greedy algorithms
in two simulated test buildings, and corroborate this finding using real sensor data 
collected in an indoor environment with multiple installed sensors. 
Our evaluation result indicates that the surrogate model of the objective function 
contains useful information in the context of assisted living. 
By leveraging this information and the inherent knowledge of the indoor space, 
DGBO achieves superb performance with respect to the activity recognition accuracy
and better sample efficiency than BO across all test environments.

Our contribution is threefold: 
1) We propose DGBO: a novel grey-box BO algorithm for optimizing sensor placement in an indoor environment;
2) We develop a simulation-based assessment and optimization framework
allowing the use of synthetic activities and movement trajectories to evaluate the black-box function,
instead of real-world traces that are difficult and expensive to collect; 
3) Through extensive experiments, we show the efficacy of DGBO in finding high-quality 
sensor placements in three different indoor environments.
Our dataset and code are available on GitHub, and we will add the link in the final version of the paper.

\section{Related Work}
\label{related_work}
\textbf{Maximizing area coverage of sensors}. 
Most related work on sensor placement seeks to maximize the area covered by the sensors, 
without considering the accuracy of downstream applications that consume the sensor data
(see for example~\cite{fanti2017smart, gungor2020respire}). 
The fundamental limitation of this approach comes from the assumption that 
all areas of the building are equally important and should be monitored 
for an application such as activity recognition.
In the real world, some building spaces are rarely occupied, 
which implies that covering the \emph{active} regions is more important. 
To address this shortcoming, several studies utilize contextual information about the building (e.g. its floor plan) to obtain a set of points that represent the potential location of occupants, 
and place sensors such that these points are maximally covered~\cite{barry2019computational,vlasenko2014smart,wu2020sensor}.
For instance, Vlasenko~\textit{et al.}~\cite{vlasenko2014smart} designed a near-optimal greedy algorithm that gets synthetic occupant trajectories
and finds a motion sensor placement for accurate indoor localization.
However, assuming all points are equally important may lead to misidentification of critical, yet rare activities, \eg bathing, in an aged-care monitoring application.
We use a greedy algorithm as a baseline.

\noindent\textbf{Maximizing the activity recognition accuracy}.
There are only a few known studies that identify the best sensor placement by maximizing the accuracy of a downstream application.
Thomas~\textit{et al.}~\cite{thomas2016genetic}
proposed a framework to find the optimal placement of omnidirectional, ceiling-mounted motion sensors
by maximizing the accuracy of an activity classifier
while minimizing the number of deployed sensors, given the occupants' activities and movement trajectories.
They used a real-world dataset to generate synthetic occupant trajectories 
and employed GA to effectively find high-quality sensor placements.
However, generating a synthetic dataset from real-world sensor readings is costly and challenging.
GA serves as the baseline for comparison.

\noindent\textbf{Sample-efficient optimization of black-box functions}.
In recent work, Deshwal~\textit{et al.}~\cite{deshwal2023bayesian} proposed a surrogate modeling approach for high-dimensional combinatorial spaces.
They used a dictionary-based embedding to map discrete structures from the input space into an ordinal feature space, allowing the use of continuous surrogate models, such as the Gaussian process.
However, the size of the search space grows exponentially with the number of possible sensor locations 
(ranging from nearly $200$ to $1000$ locations in a small, \eg 700-square-foot, indoor environment). 
Due to its inability to directly control the sensor numbers, our assessments show that applying this method to the sensor placement problem often leads to installing ${>}500$ sensors in our environments, 
which is overly costly, rendering this approach of limited practical value in this domain.

\noindent\textbf{Grey-box Bayesian optimization}.
Some studies have explored using the domain-specific knowledge available about $f(x)$, thereby
treating $f(x)$ as a grey-box function~\cite{astudillo2021thinking}. 
They have sought a method to capture this knowledge and
incorporate it into the definition of an \textit{acquisition function}, 
which decides on the next query point in the optimization process. 
For example, Weissteiner~\textit{et al.}~\cite{weissteiner2023bayesian}
proposed a method for estimating an upper uncertainty bound, 
to design a new acquisition function for BO-based combinatorial auctions optimization.

\noindent\textbf{Novelty of this work}. 
We cast sensor placement in an indoor environment as a Bayesian optimization problem.
We then introduce a novel grey-box BO, called DGBO, 
and compare the resulting motion sensor placement with those found by BO, GA, and greedy.
We show that the available domain-specific knowledge in our problem guides DGBO towards 
better sensor placements, and increases its sample efficiency.

\section{Methodology}
\label{methodology}

\begin{figure}[t!]
    \centering
    \centerline{\includegraphics[width=0.6  \columnwidth]{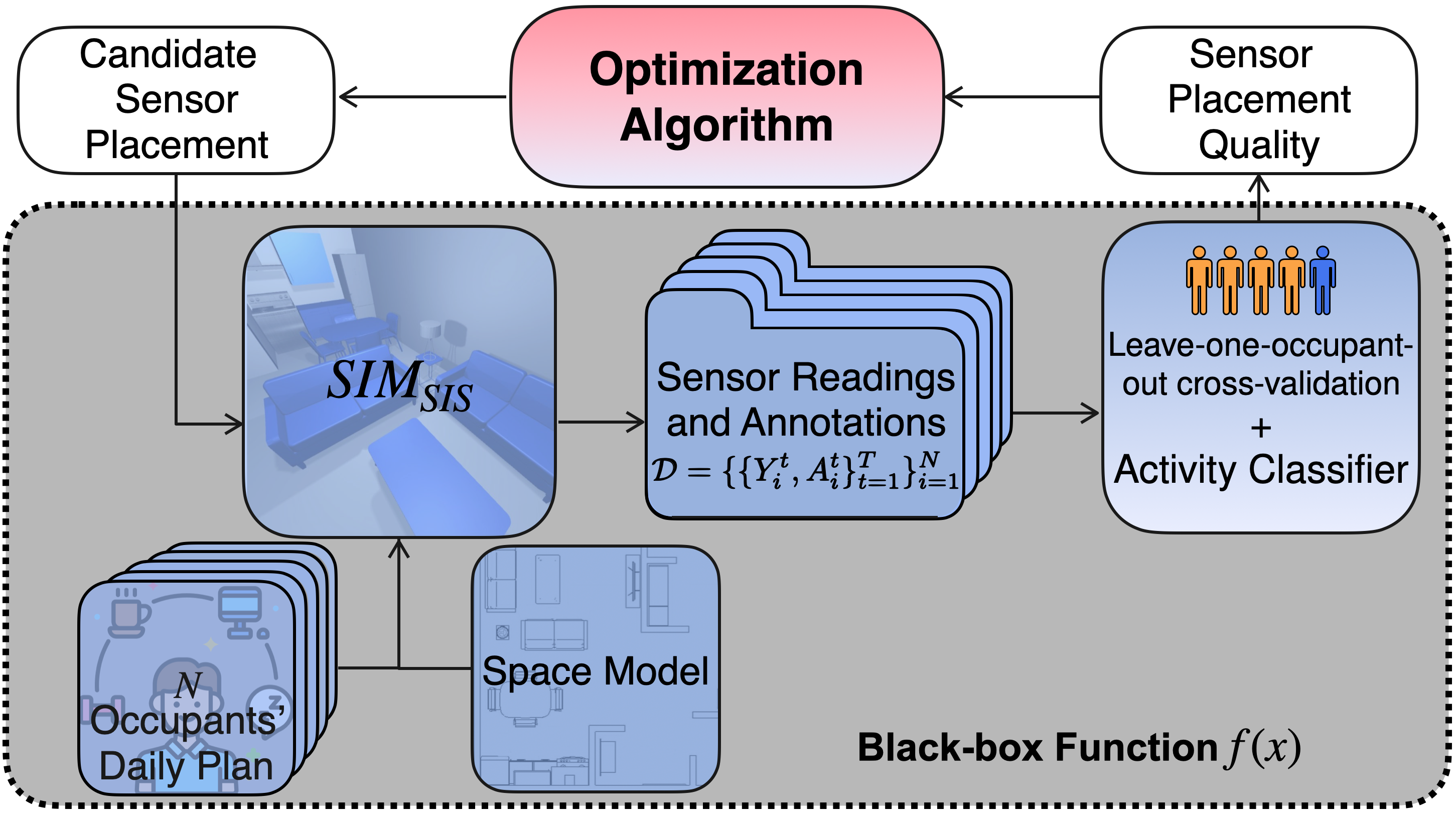}}
    \caption{An overview of the proposed optimization framework for sensor placement.}
    \label{fig_framework}
\end{figure}

We present our simulation-driven motion sensor placement framework in Figure~\ref{fig_framework}.
At the core of this framework lies an optimization algorithm
that finds a sensor placement maximizing the performance of a model 
that classifies \emph{activities of daily living}~(ADL), given data generated by the sensors.
To efficiently obtain motion sensor data for each candidate sensor placement, 
we use a simulator of smart indoor spaces (\simsis) developed by Golestan~\textit{et al.}~\cite{golestan2020towards} (such simulators are thoroughly studied~\cite{briscoe2022adversarial,dahmen2019synsys}).
the \simsis~gets ADL plans of
$N$ occupants and the location of motion sensors,
and simulates motion sensor triggers.
The sensor triggers and corresponding activities are then passed 
to the activity classifier module to obtain the result.
This module trains an ADL classifier using $80\%$ of the dataset, 
and evaluates its performance on the remaining $20\%$ of the dataset (test set). 
Putting it all together, we treat the simulation of the indoor environment as a stochastic function, $f(x)$,
that gets as input a candidate sensor placement $x$ and outputs a noisy observation,
which is a measure of performance of an activity recognition model. 

\subsection{Simulator of Smart Indoor Spaces}
The \simsis~is an open-source\footnote{https://github.com/shadangolestan/SIM{\textunderscore}SIS-Simulator} indoor environment simulator capable of realistically modeling occupants' behavior and 
simulating motion sensor readings.
It receives a space model, \ie the specification of an indoor environment, daily plans of $N$ occupants \ie regular ADLs, such as sleeping and cooking, and a candidate sensor placement which describes the number and locations of motion sensors. 
Each sample from the model generates a slightly different daily plan in terms of ADL order and duration.
The \simsis~generates a synthetic dataset with sensor readings and 
their corresponding activity for each of the $N$ occupants:
\begin{equation}
\label{simsis_output}
\mathcal{D} = \{\{Y^t_i, A^t_i\}^T_{t=1} \}^{N}_{i=1}
\end{equation}
\noindent where $Y^t_i$ and $A^t_i$ denote respectively the 
sensor triggers (a binary vector of size $D$ where $D$ is the number of sensors installed in the environment) and the activity of the $i$-th occupant at time $t$. 
An ideal sensor placement\footnote{We assume the motion sensors are omnidirectional and attached to the ceiling, so the sensor placement problem reduces to determining the location of each sensor in a 2D space.} should result in a dataset where each activity has a distinct enough pattern 
in the space of sensor readings so that an activity recognition model can recognize it~\cite{freedman2019unifying}.

\subsection{Activity Classifier}
The activity classifier gets the dataset $\mathcal{D}$ generated by a fixed set of sensors installed in the environment and applies the leave-one-out cross-validation method. 
Specifically, it considers the data that pertains to one occupant 
as the test set and the data of the remaining occupants as the training set. 
This process is repeated for each occupant to obtain different test sets.
Following Cook~\textit{et al.}~\cite{cook2012casas}, 
we use a random forest as the activity recognition model.
For each test set, the activity classifier calculates the macro-averaged F1-score of the ADL 
and then outputs the arithmetic mean of the F1-scores (given below) as the model performance:
\begin{equation}
\label{metric}
    F^1 = \frac{1}{N}\Sigma_{i = 1}^N{\frac{1}{M}\Sigma_{j = 1}^M\frac{tp_j}{tp_j + \frac{1}{2} ( fp_j + fn_j )}}
\end{equation}
\noindent Here $N$ is the number of occupants, $M$ is the number of activities, and $tp_j$, $fp_j$, and $fn_j$ are true positive, false positive and false negative of the \textit{j}-th activity, respectively. 
Notice that $F^1$ is sensitive to both false negative and false positive. 
We argue that it is a good performance measure, because both of these factors are important in the activity recognition task, and in particular for elderly monitoring systems 
where failing to alert the caregiver or issuing a false alarm could have dire consequences. 
Equation~\ref{metric} calculates the macro-averaged F1-score over the $M$ activities,
since there are rare and short, but precarious activities, \eg bathing, 
that should be deemed as important as other activities in our problem.

\subsection{Optimization Algorithm}
The optimization algorithm module receives an observation, which is the performance of the activity recognition model expressed in Equation~\ref{metric} ($f(x){=}F^1$), 
and proposes a candidate sensor placement in each iteration to maximize this performance measure. 
This optimization can be performed using vanilla BO, DGBO (our proposed method), greedy and genetic algorithms, 
which are two popular heuristic search algorithms adopted in the literature.

\subsubsection{Problem Formulation}
Let us denote the possible sensor locations in the indoor environment as a set $\mathcal{L}$
define as:
\begin{equation}
\label{grid}
    \mathcal{L}=\{l_i\}_{i=1}^{L} 
\end{equation}
We seek to find the subset of $\mathcal{L}$
such that if motion sensors are installed at these locations 
then $f(x)$ will be maximized.

\subsubsection{Bayesian Optimization}
BO is a sequential search strategy for optimizing a black-box function. 
In our problem, this function is $f$ which returns the performance of the activity recognition model. 
Thus, BO solves the following problem:
\begin{equation}
\label{f}
    \max_{x \in S} {f(x)}
\end{equation}
\noindent \sloppy where $S$ is the 
search space containing all possible placements of $D$ sensors $(1{\leq}D{\leq}L)$
and $x$
represents a feasible sensor placement. 
Note that the size of $S$ grows exponentially with the number of sensors: $|S|{=}\binom{L}{D}$. 
The BO assumes that observations of $f(x)$ are the results of a stochastic process and are independent.
There are two sources of noise in $f$: 
1)~\simsis:
it does not necessarily capture all dynamics and uncertainties that exist in the real-world;
2)~the activity classifier: the random forest fits several decision tree classifiers on different randomly generated sub-samples of the dataset. Thus, each execution might yield slightly different results. 
BO approximates $f$ using a surrogate model, denoted $\hat{f}$, 
given a set of observations $\mathcal{Z}$:
\begin{equation}
\label{surrogate_function}
    \hat{f}(x) = p(f(x) | \mathcal{Z}).
\end{equation}
According to this definition, the surrogate model $\hat{f}$ estimates $F^1$ (Equation~\ref{metric}) given a sensor placement $x$. 
We use the Probabilistic Random Forest (PRF) as the BO's surrogate model~\cite{hutter2011sequential}. 
We use the observations to compute the mean and standard deviation of
$\hat{f}(x)$ at a sensor placement $x$ and denote them as $\mu_x$ and $\sigma_x$.
Hence, we can write $\hat{f}(x){=}\mu_x{+}\sigma_x u$ where $u{\sim}\mathcal{N}(0,1)$.
In each iteration $n$, the surrogate model is indeed a prior over $f$ given observations received 
from the first iteration to the $n$-th iteration: $\mathcal{Z}_{1:n}{=}\{ (x_i, f(x_i) \}_{i=1}^n$. 

The surrogate model $\hat{f}$ is used to form an acquisition function $\alpha(x; \hat{f})$, 
which determines the next candidate sensor placement ($x_{n{+}1}$) to evaluate in iteration $n{+}1$. 
We use the
expected improvement (EI) function~\cite{frazier2018tutorial}:
\begin{equation}
\label{acquisition_function}
\begin{split}
    &\alpha_{\text{EI}}(x; \hat{f}) = \int\limits_{-\infty}^{\infty} \max\big(\hat{f}(x) - f(x^*), 0\big) \phi(u) \,du =\\
                       & (\mu_x\!-\!f(x^*))\Phi(\frac{\mu_x\!-\!f(x^*)}{\sigma_x}) + \sigma_x \phi(\frac{\mu_x\!-\!f(x^*)}{\sigma_x})
\end{split}
\end{equation}
where $f(x^*){=}\max\{f(x_1), ..., f(x_n) \}$ is the observation that corresponds to the best sensor placement found so far $x^*$ (i.e. the incumbent);
 $\Phi(.)$ and $\phi(.)$ denote the cumulative density function~(CDF) and 
probability density function~(PDF) of the standard normal distribution, respectively. 
The acquisition function is used to find potentially better sensor placements. 
The next query point is given by: 
\begin{equation}
\label{acq_deployment}
    x_{n+1} = \argmax_x{\alpha_{\text{EI}}(x; \hat{f})},
\end{equation} which is the point with the largest expected improvement. 
The choice of EI is due to its widespread use
and because it strikes a good balance between exploration and exploitation. 

After evaluating $x_{n+1}$, the corresponding observation,
\ie ${z_{n+1}{=}(x_{n+1}, f(x_{n+1}))}$,
is appended to the sequence of past observations, $\mathcal{Z}_{1:n}$, 
to obtain $Z_{1:n+1}{=}  \{Z_{1:n}, z_{n+1} \}$. 
Next, $\mathcal{Z}_{1:n+1}$ is used to update the surrogate model $\hat{f}$, 
resulting in the posterior surrogate model that better approximates $f$. 
The posterior surrogate model is then used as a prior for obtaining $x_{n+2}$. 
The process continues for $1000$ iterations and the best sensor placement found is reported. 
We randomly choose the initial placement ($x_1$) and implement BO
using the package from~\cite{li2021openbox}.

\subsubsection{Distribution-Guided Bayesian Optimization (DGBO)}
We now describe the proposed grey-box BO.
Similar to BO, DGBO utilizes a surrogate model and an acquisition function.
It also takes advantage of the distribution of indoor activities, 
which can be captured from the evaluation of $f(x)$.

For each possible sensor location $l_i{\in}\mathcal{L}$,
we define an activation region $R_{i}$ which contains every point that would be within the range of 
a motion sensor installed at $l_i$.
Note that the activation regions of two sensors may overlap.
The objective is to estimate an unknown function $I(R_{i})$ that outputs the amount of information $R_{i}$ provides.
This function provides insight into how useful a sensor is when monitoring a region.
Intuitively, this should depend on the spatial distribution of activities and 
the location of other sensors in the environment.
To obtain prior information about each activation region $R_{i}$,
we query $f(x)$ at $l_i$; this is equivalent to installing exactly 
one sensor in the middle of $R_i$ and using only its data for activity recognition.
DGBO uses the prior information about each activation region $R_{i}$ 
to measure the amount of information that this region could provide at iteration $n$,
and inserts it into a tabular data structure,
denoted $\mathcal{I}_{0:n}\{R_i\}$ and marked \textit{Information Profiles} in Figure~\ref{intro_fig}.
Thus, after receiving the $n$-th observation ($z_n{=}(x_n,f(x_n))$),
we update $\mathcal{I}_{0:n}\{R_{i}\}$ for every $R_{i}$ that the respective $l_i$
represents the location of a sensor in $x_n$ as follows:
\begin{equation}
\label{info}
\mathcal{I}_{0:n}\{R_i\} = \left\{\mathcal{I}_{0:n-1}\{R_i\}, \frac{\mathcal{I}_{0:0}\{R_i\}}{\Sigma_{l_j \in x_n}{\mathcal{I}_{0:0}\{R_j\}}}f(x_n)\right\}
\end{equation} where $\mathcal{I}_{0:0}\{R_i\}$ is the prior information of region $R_i$. Equation~\ref{info} contributes a portion of $f(x_n)$ to $R_i$ 
based on its prior information relative to other regions' prior information.
This is similar to the temporal credit assignment problem~\cite{sutton1984temporal} in Reinforcement Learning, wherein a sequence of actions taken by an agent 
receive credit based on its outcome. 
With this analogy, our approach, which is a form of spatial credit assignment, 
identifies the contribution of a number of regions 
to the particular level of performance achieved as a result of installing sensors 
in the middle of these regions.
We use Equation~\ref{info} to build a prior over $I(R_{i})$:
\begin{equation}
\label{surrogate_function_I}
    \hat{I}(R_{i}) = p(I(R_{i})~|~\mathcal{I}_{0:n}\{R_i\}).
\end{equation}
Assuming $\hat{I}(R_{i})$ follows a Gaussian distribution ($\mathcal{N}(\mu_{R_{i}}, \sigma_{R_{i}})$), 
we can use the mean and standard deviation of $\mathcal{I}_{0:n}\{R_i\}$ to calculate its parameters.
We define the relative \emph{information gain} of a region compared to the information gain of the incumbent:
\begin{equation}
\label{info_gain}
    I^{+}_n(R_{i}){=}\max(\hat{I}(R_{i}) - I^{*}, 0)
\end{equation} 
where $I^{*}{=}\frac{1}{D} \sum_{l_j \in x^*} {I}^{+}_{n-1}(R_j)$ is the average information gain of regions in incumbent sensor placement $x^*$, and $I^{+}_0(R_{i}){=}\mathcal{I}_{0:0}\{R_{i}\}$. 
The expected information gain of each region is then calculated by taking 
the expected value of Equation~\ref{info_gain} (see module~2 in Figure~\ref{intro_fig}).

For each placement $x$, our acquisition function, $\alpha_{\text{DG}}(x)$, calculates 
the average expected information gain of the regions $R_i$ where a sensor is placed in $x$:
\begin{equation}
    \alpha_{\text{DG}}(x;\hat{f}) = \tfrac{1}{D} \sum_{l_i \in x} {\mathbb{E}[I_n^{+}(R_{i})]} 
\end{equation} 
With a reparameterization,
we can write $\hat{I}(R_{i}){=}\mu_{R_{i}}{+}\sigma_{R_{i}}u$ where $u{\sim}\mathcal{N}(0,1)$:
\begin{align*}
\nonumber &\alpha_{\text{DG}}(x;\hat{f}){=}\tfrac{1}{D} \sum_{l_i \in x} \int\limits_{u_0}^{+\infty}(\mu_{R_{i}}{+}\sigma_{R_{i}}u{-}I^{*}) \phi(u) \, du \\ 
\nonumber &{=}\tfrac{1}{D} \sum_{l_i \in x} \left(~\int\limits_{u_0}^{+\infty} (\mu_{R_{i}}{-}I^{*})\phi(u) \, du + \int\limits_{u_0}^{+\infty}\sigma_{R_{i}}u\phi(u) \, du \right)
\end{align*}
\begin{align}
\nonumber &{=}\tfrac{1}{D} \sum_{l_i \in x}\left((\mu_{R_{i}}{-}I^{*})\int\limits_{u_0}^{+\infty}\phi(u) \, du + \frac{\sigma_{R_{i}}}{\sqrt{2\pi}}\int\limits_{u_0}^{+\infty}ue^{\tfrac{-u^2}{2}}\, du \right) \\
\nonumber &{=}\tfrac{1}{D} \sum_{l_i \in x}\left((\mu_{R_{i}}{-}I^{*})(1{-}\Phi(u_0)) - \frac{\sigma_{R_{i}}}{\sqrt{2\pi}}\left[e^{\tfrac{-u^2}{2}}\right]_{u_0}^{+\infty} \right)\\
&{=}\tfrac{1}{D} \sum_{l_i \in x} (\mu_{R_{i}} - I^{*})\Phi(\frac{\mu_{R_{i}}{-}I^{*}}{\sigma_{R_{i}}}) + \sigma^2_{R_{i}}\phi(\frac{\mu_{R_{i}}{-}I^{*}}{\sigma_{R_{i}}})
\end{align}
where $u_0=\frac{I^{*}{-}\mu_{R_{i}}}{\sigma_{R_{i}}}$ 
is the value of $u$ at which the information gain becomes zero. 
To exploit the exploitation and exploration tendency of $\alpha_{\text{DG}}$ and $\alpha_{\text{EI}}$, respectively, we use scalarization and maximize $\alpha_{\text{DG}}(x;\hat{f}){+}\alpha_{\text{EI}}(x;\hat{f})$ 
to choose the next query point (see module~3 in Figure~\ref{intro_fig}).
Intuitively, this point maximizes the sum of expected improvement and expected information gain,
rather than any of them.

\section{Experiments}
\label{results}
\subsection{Simulation Case Study}
\begin{figure}[tb]
     \centering
     \begin{subfigure}{0.3\columnwidth}
         \centering
         \centerline{\includegraphics[width=\columnwidth]{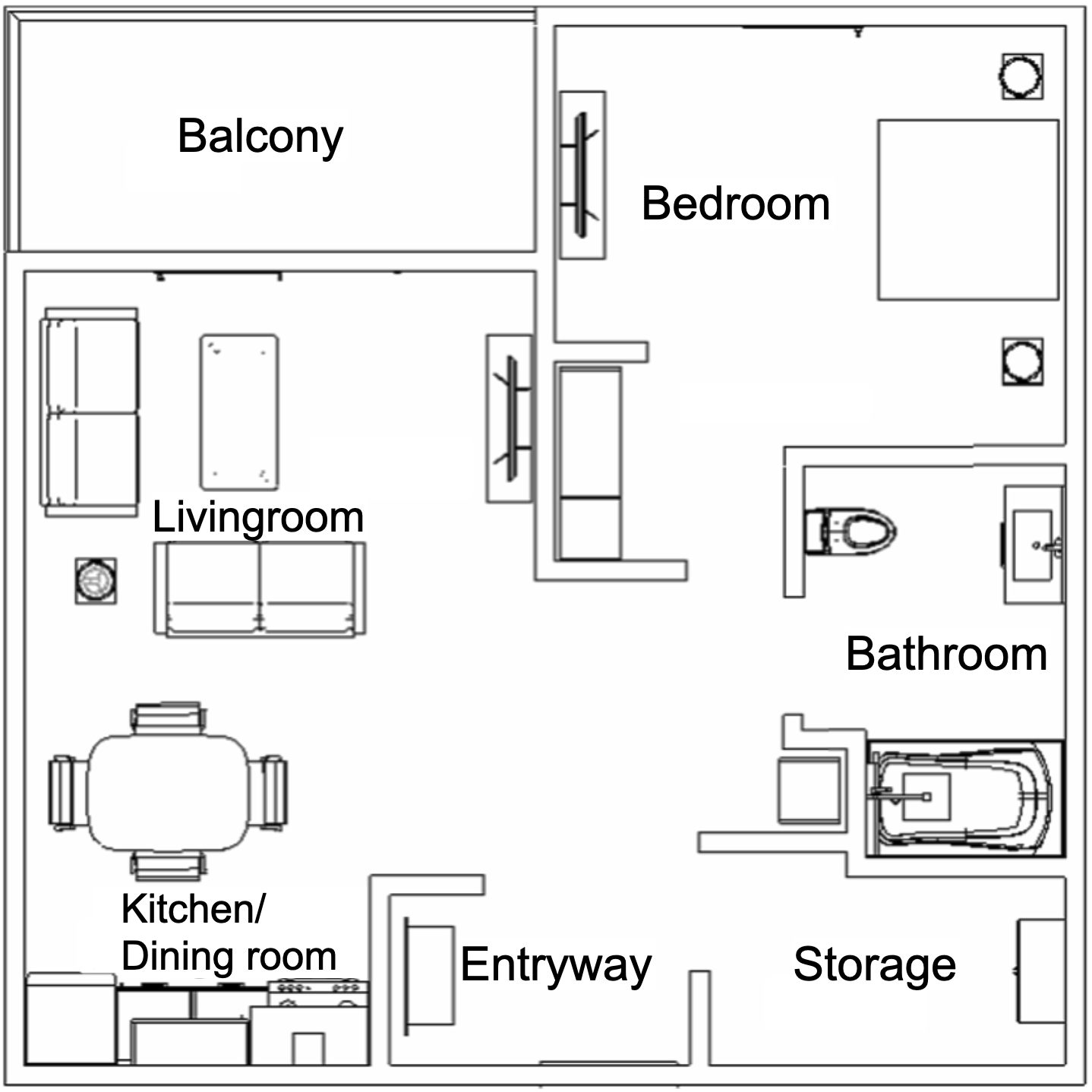}}
         \caption{Testbed 1 (T1).}
         \label{TB1}
     \end{subfigure}
     \hspace{0.5em}
     \begin{subfigure}{0.31\columnwidth}
         \centering
         \centerline{\includegraphics[width=\columnwidth]{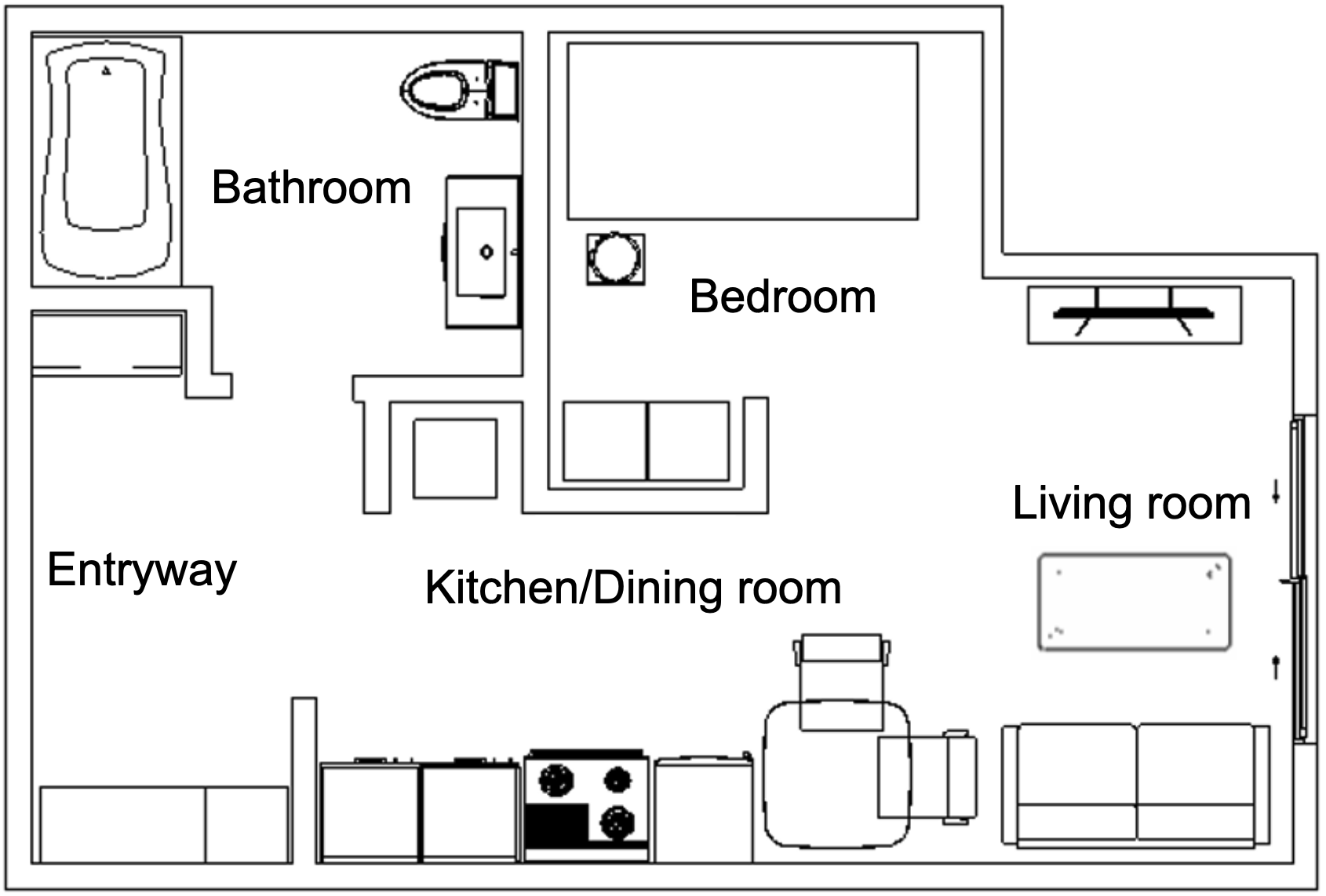}}
         \caption{Testbed 2 (T2).}
         \label{TB2}
     \end{subfigure}
     \caption{The floor plan/space model of the two apartments: 
     (a) an $8{\times}8(m^2)$ one-bedroom suite; (b) a $5.2{\times}8(m^2)$ studio.}
     \label{testbed}
\end{figure}

We consider two testbeds for our simulation case study: an $8{\times}8~(m^2)$ fully-furnished, one-bedroom suite (marked T1) and a $5.2{\times}8~(m^2)$ fully-furnished, studio suite (marked T2) from the Lifestyle Options Retirement Communities~\cite{lifestyleoptions}. 
Figure~\ref{testbed} shows the floor plan of these suites.
The suites are designated for older adults needing independent living, assisted living, or memory care. 
We choose these suites as our testbeds due to their differences, 
such as layout and size, which could lead to a diverse set of traces and unique challenges. 
Specifically, T2 exhibits distinct movement trajectories in some areas, \eg the kitchen, 
since the entryway and bathroom are accessible only from the kitchen,
and T1 exhibits slightly scattered movement trajectories and has no trajectories in the balcony. 

Similar to most related work, such as \cite{thomas2016genetic},
we postulate that the set of possible sensor locations forms a 2D grid on the given floor plan. 
Thus, ${L}{=}{H}{\times}{W}$ (see Equation~\ref{grid}) 
with ${H}{=}\ceil{h/\epsilon}{-}1$ and ${W}{=}\ceil*{w/\epsilon}{-}1$ 
where $h$ and $w$ are the height and width of the indoor environment, respectively, 
and $\epsilon$ is the spacing between consecutive rows and columns. 
Higher $\epsilon$ values indicate lower granularity and lower computation overhead.

The case study includes five occupants (${N}{=}5$) 
performing various ADLs independently in each testbed. 
These activities and corresponding sensor triggers are simulated by \simsis.
Table~\ref{tabADLs} shows an ordered list of $23$ detailed activities 
the occupants perform in $196$ minutes in total (in Equation~\ref{simsis_output}, $A$ is the list of detailed activities and ${T}{=}\frac{196{\times}60}{3}$ since data is collected every $3$ seconds). 
The order of performing ADLs and some detailed activities in each ADL
(denoted using the same superscript in Table~\ref{tabADLs}) 
can be interchanged, producing slightly different activity plans.

\begin{table*}[tb]
\footnotesize
\centering
\begin{tabular}{|p{2.0cm}|p{11.0cm}|}
\hline
 \textbf{ADL} & \textbf{Seq. of detailed activities (duration in minutes)} \\ 
\hline

                      Bathing & 
                      Undress (5), Take a shower (15), Dress (6)\\  \hline
                      
                      Hygiene & 
                      Use toilet (3), Wash hands (3) \\ \hline
                      
                      Dining routine
                      & 
                      Make tea (10), Grab ingredients (2), Fry eggs (10), Toast breads (5), Grab utensils (1), Eat (10), Take medicine (2), Wipe dining table$^a$ (5), Wash dishes$^a$ (3), Clean kitchen$^a$ (5) \\ \hline
                      
                      Brooming & 
                      Grab the broom from storage (2), Broom (7), Return the broom (2) \\ \hline
                      
                      Others & 
                      Sit and work with tablet$^b$ (30), Exercise$^b$ (30), Watch TV$^b$ (15), Iron$^b$ (5), Sleep$^b$ (20) \\ 
                      
                      \hline
\end{tabular}
\caption{ADLs in our simulated case study. ADLs and detailed activities with the same superscript can be shuffled.}
\label{tabADLs}
\end{table*}

\begin{table*}[tb]
\footnotesize
\centering
\begin{tabular}{|p{0.3cm}|p{1.9cm}|p{1.9cm}|p{1.9cm}|p{1.9cm}|p{1.9cm}|p{1.85cm}|}
\hline
    
    \multicolumn{2}{|c|}{\multirow{2}{*}{\textbf{Testbed}}}&
    \multicolumn{4}{c|}{\textbf{Avg. $F^1$ ${\pm}$ one standard deviation (no. sensors used)}} \\ \cline{3-6}
    
    \multicolumn{2}{|c|}{ }
    &
    \textbf{GA} &
    \textbf{Greedy} &
    \textbf{BO} &
    \textbf{DGBO} \\ \hline

    \multirow{3}{*}{\textbf{T1}} & 
    $\mathbf{\epsilon{=}0.25(m)}$ &
    $56.7{\pm}1.0$ (9) &
    --- &
    $76.9{\pm}0.1$ (9) &
    $77.6{\pm}1.1$ (9) \\

     &
    $\mathbf{\epsilon{=}0.5(m)}$ &
    $59.7{\pm}0.4$ (11) &
    --- &
    $75.6{\pm}2.0$ (7) &
    $77.5{\pm}0.1$ (15) \\

     &
    $\mathbf{\epsilon{=}1.0(m)}$ &
    $54.5{\pm}1.0$ (9) &
    $69.7{\pm}1.1$ (13) &
    $75.0{\pm}1.1$ (9) &
    $77.6{\pm}0.2$ (9) \\ \cline{2-6}

    \multirow{3}{*}{\textbf{T2}} & 
    $\mathbf{\epsilon{=}0.25(m)}$ &
    $42.9{\pm}1.5$ (10) &
    --- &
    $72.3{\pm}1.6$ (11) &
    $72.3{\pm}1.8$ (11) \\

     &
    $\mathbf{\epsilon{=}0.5(m)}$ &
    $42.7{\pm}0.4$ (6) &
    $59.8{\pm}2.7$ (7) &
    $67.7{\pm}0.3$ (11) &
    $68.7{\pm}2.6$ (15) \\

     &
    $\mathbf{\epsilon{=}1.0(m)}$ &
    $40.7{\pm}1.3$ (5) &
    $66.8{\pm}1.3$ (15) &
    $66.9{\pm}1.1$ (13) &
    $69.6{\pm}0.4$ (11) \\ \hline

    \multicolumn{2}{|l|}{\textbf{Aruba}} & 
    $60.2{\pm}1.2$ (7) &
    $64.0{\pm}1.3$ (15) &
    $75.7{\pm}0.2$ (9) &
    $76.3{\pm}0.2$ (9) \\
        
\hline
\end{tabular}
\caption{The performance of GA, Greedy, BO and DGBO in terms of the macro-averaged $F^1$ (Equation~\ref{metric}) in T1, T2 and Aruba.
A dash indicates that no sensor placement was found after $1000$ queries. We report the median number of sensors for GA.}
\label{resultsTable}
\end{table*}

\begin{figure*}[tb]
    \centering
     \begin{subfigure}{0.32\columnwidth}
         \includegraphics[width=\columnwidth]{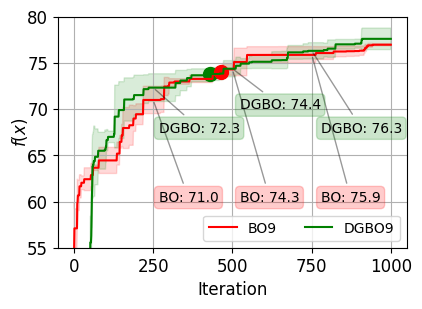}
         \caption{$\epsilon = 0.25$ (T1)}
         \label{T1_25_}
     \end{subfigure}
     \begin{subfigure}{0.32\columnwidth}
         \includegraphics[width=\columnwidth]{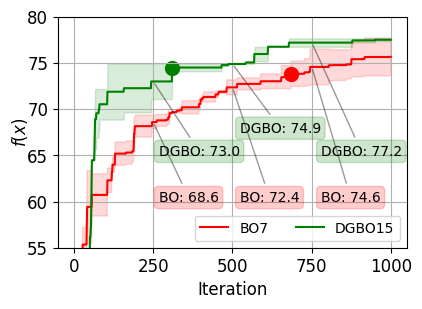}
         \caption{$\epsilon = 0.5$ (T1)}
         \label{T1_5_}
     \end{subfigure}
     \begin{subfigure}{0.32\columnwidth}
         \includegraphics[width=\columnwidth]{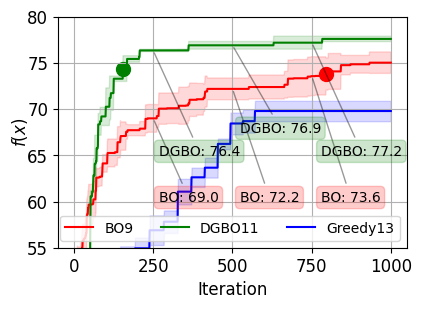}
         \caption{$\epsilon = 1$ (T1)}
         \label{T1_1_}
     \end{subfigure}
     
     \begin{subfigure}{0.32\columnwidth}
         \includegraphics[width=\columnwidth]{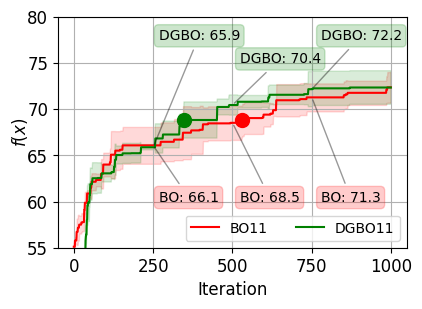}
         \caption{$\epsilon = 0.25$ (T2)}
     \label{T2_25_}
     \end{subfigure}
     \begin{subfigure}{0.32\columnwidth}
         \includegraphics[width=\columnwidth]{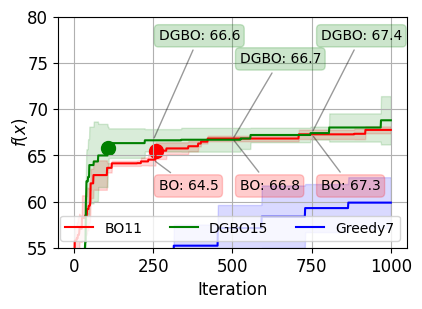}
         \caption{$\epsilon = 0.5$ (T2)}
     \label{T2_5_}
     \end{subfigure}
     \begin{subfigure}{0.32\columnwidth}
         \includegraphics[width=\columnwidth]{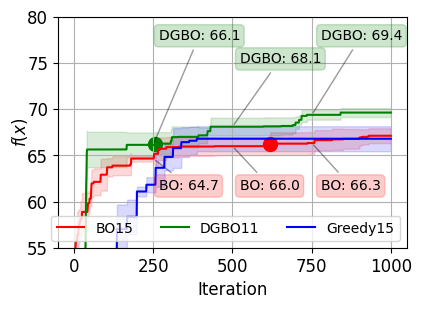}
         \caption{$\epsilon = 1$ (T2)}
     \label{T2_1_}
     \end{subfigure}
     
     \begin{subfigure}{0.32\columnwidth}
         \includegraphics[width=\columnwidth]{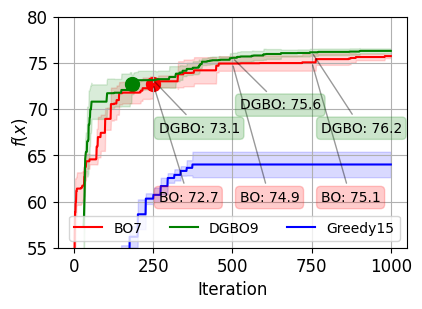}
         \caption{Aruba (real-world dataset)}
     \label{Aruba_}
     \end{subfigure}
     \caption{The average performance of DGBO, BO, and greedy (when available) for the best number of sensors found by each method versus the number of $f(x)$ queries.}
     \label{average_performance}
\end{figure*}

\subsection{Real-World Case Study}
We use a publicly available, real-world dataset called Aruba~\cite{cook2012casas}. 
This dataset was collected from the home of an adult who lives alone but has regular visitors. 
The dataset contains $219$ days worth of data generated by $31$ motion sensors installed in that home. 
The following $11$ ADLs were manually labeled: 
1) meal preparation,
2) relax,
3) eating,
4) work,
5) sleeping,
6) wash dishes,
7) bed to toilet,
8) enter home,
9) leave home,
10) housekeeping, and
11) respirate.
In this case, \simsis~gets as input the real-world dataset 
and instead of simulating human activities and corresponding sensor triggers, 
it simply filters a subset of the sensor data based on the sensors 
present in the candidate sensor placement.
This portion of the dataset is then used for activity recognition, 
with the first $70\%$ of the days comprising the training set and the rest comprising the test data.

\subsection{Results}
We compare motion sensor placements found by DGBO (the proposed method), BO, GA, and greedy algorithm
in our case studies. 
The implementation of our baselines, i.e. GA and greedy algorithm, is described in the appendix.
For each value of $\epsilon{\in}\{0.25m, 0.5m, 1.0m\}$ in the simulation case study,
we run each algorithm $5$ times, using different seeds. 
Given the range of motion sensors in the simulator (a circle with radius of $1$ meter), 
these $\epsilon$ values are reasonable because they allow some overlap 
between the areas covered by multiple sensors 
and maintaining a clear line of sight despite the obstacles that exist in the environment.
For greedy, BO, and DGBO,
we repeat the process after setting the total number of sensors to 
$5$, $7$, $9$, $11$, $13$, and $15$. 
GA decides the number of sensors automatically.
Each algorithm can query the black-box function 1,000 times. 
Table~\ref{resultsTable} shows the performance (Equation~\ref{metric}) of DGBO, BO, greedy and GA
for the number of sensors that led to the best performance in each case (mentioned in brackets).
In both T1 and T2, the greedy algorithm mostly exhausts the 1,000 black-box function queries 
for $\epsilon{=}0.25$ and $0.5$, failing to find a solution.
For example, when $\epsilon{=}0.5$ in T1, 
$1115$ function queries would be needed to place $5$ sensors (see appendix for details).

We make the following observations: 
1) Both DGBO and BO significantly outperform GA and greedy for all $\epsilon$ values.\footnote{For each $\epsilon$ value, a two-tailed t-test is used to decide if there exists a significant difference ($p{<}0.05$).}
This confirms our hypothesis that the surrogate model of $f(x)$ contains useful information in this context.
2) DGBO consistently attains better performance than BO in all testbeds with any $\epsilon$ value. 
3) Greedy algorithm performs well in T2 with $\epsilon{=}1$. 
We attribute this to the rather small search space for this value of $\epsilon$,
increasing the chance of reaching the global optimum.
4) DGBO performs robustly across all testbeds, 
as evidenced by its convergence to a similar level of performance compared to the other methods.
We believe this is because the information profiles guide DGBO towards 
more interesting parts of the indoor environment (see Figure~\ref{intro_fig}).

We investigate the fourth observation in detail.
First, Figure~\ref{average_performance} depicts the average performance of DGBO, BO and greedy 
across the 5~runs after each iteration. 
We specifically focus on the best-performing number of sensors for each algorithm.
We witness that DGBO quickly finds a high-quality sensor placement in all testbeds. 
To compare the sample efficiency of DGBO and BO, 
we find the first iteration at which DGBO and BO
reach the $95\%$ confidence interval (CI) of the best performance of DGBO after 1,000 iterations;
these iterations are marked with green and red dots in Figure~\ref{average_performance}, respectively.
Table~\ref{conv} shows how many fewer queries are required by DGBO 
to attain the same performance as BO, as the percentage of the number of queries executed by BO.
On average, DGBO requires $55.4\%$, $58.9\%$, and $39\%$ fewer queries than BO in T1, T2, and Aruba, respectively.
This is a significant improvement, especially because these queries are typically expensive.

\begin{figure}[tb]
    \centering
     \begin{subfigure}{0.2445\columnwidth}
         \centering
         \centerline{\includegraphics[width=\columnwidth]{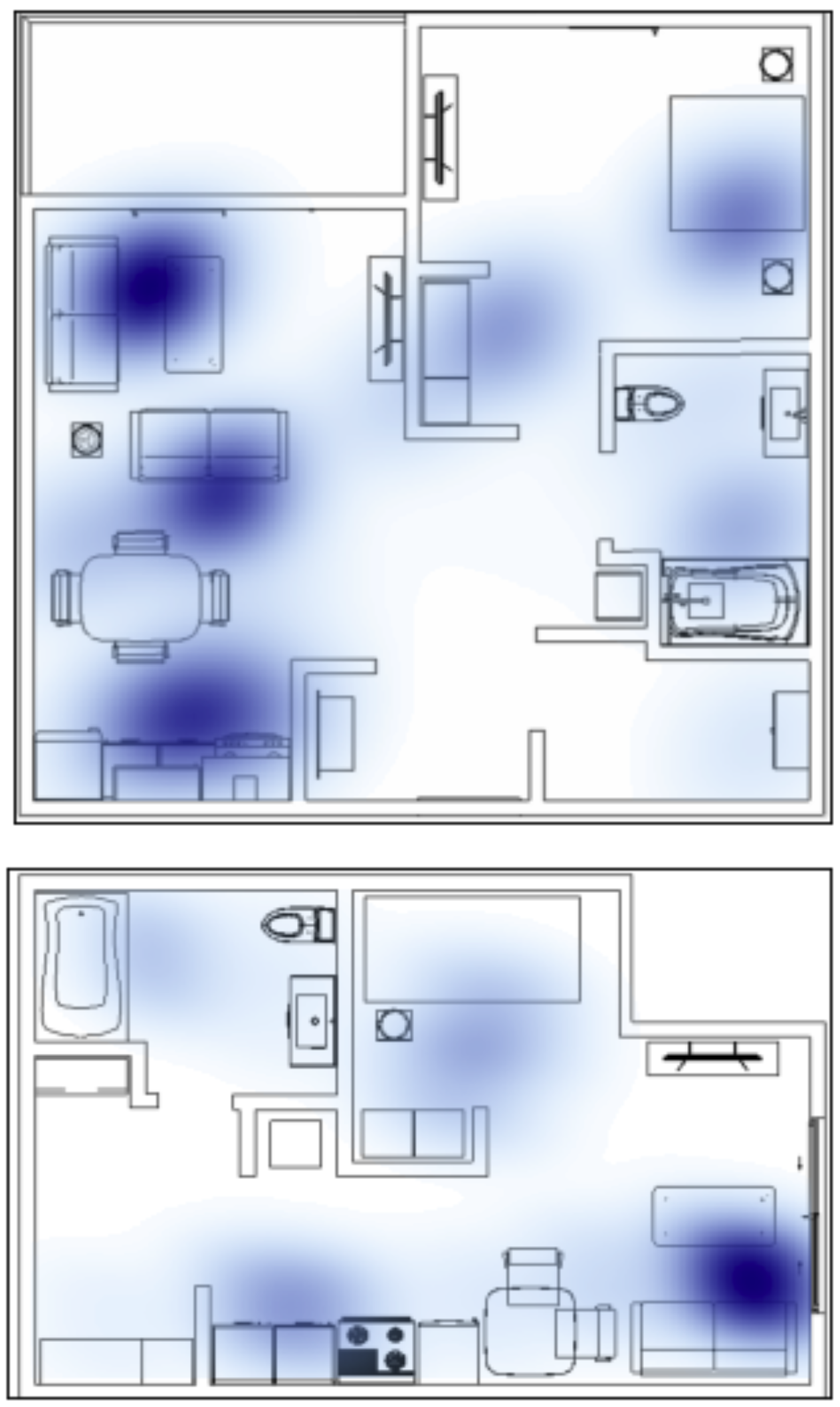}}
         \caption{}
         \label{activitiesDistribution}
     \end{subfigure}
      \begin{subfigure}{0.241\columnwidth}
         \centering
         \centerline{\includegraphics[width=\columnwidth]{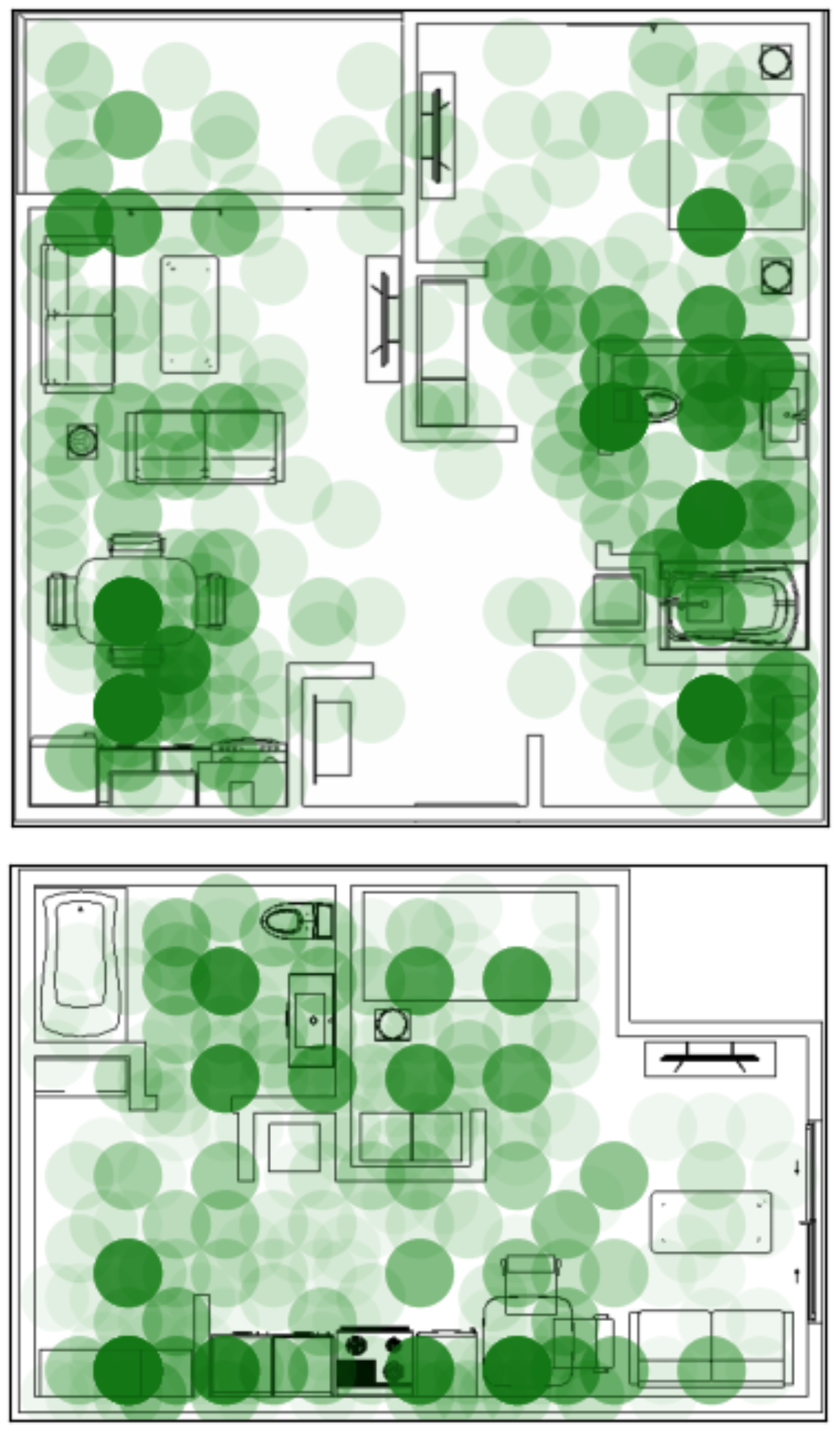}}
         \caption{}
         \label{DGBOlocations}
     \end{subfigure}
      \begin{subfigure}{0.24\columnwidth}
         \centering
         \centerline{\includegraphics[width=\columnwidth]{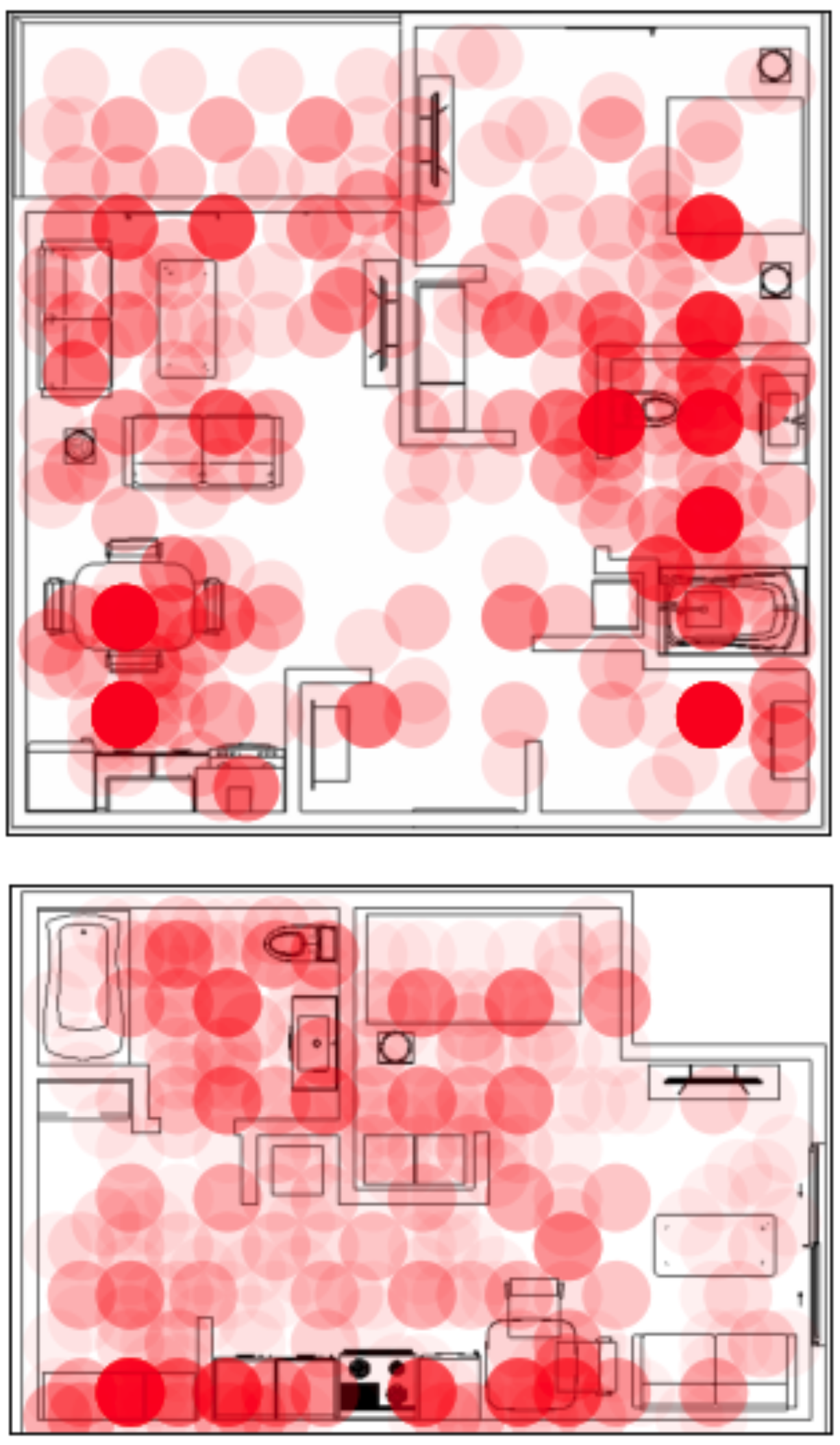}}
         \caption{}
         \label{BOlocations}
     \end{subfigure}
     \begin{subfigure}{0.08\columnwidth}
         \centering
         \centerline{\includegraphics[width=\columnwidth]{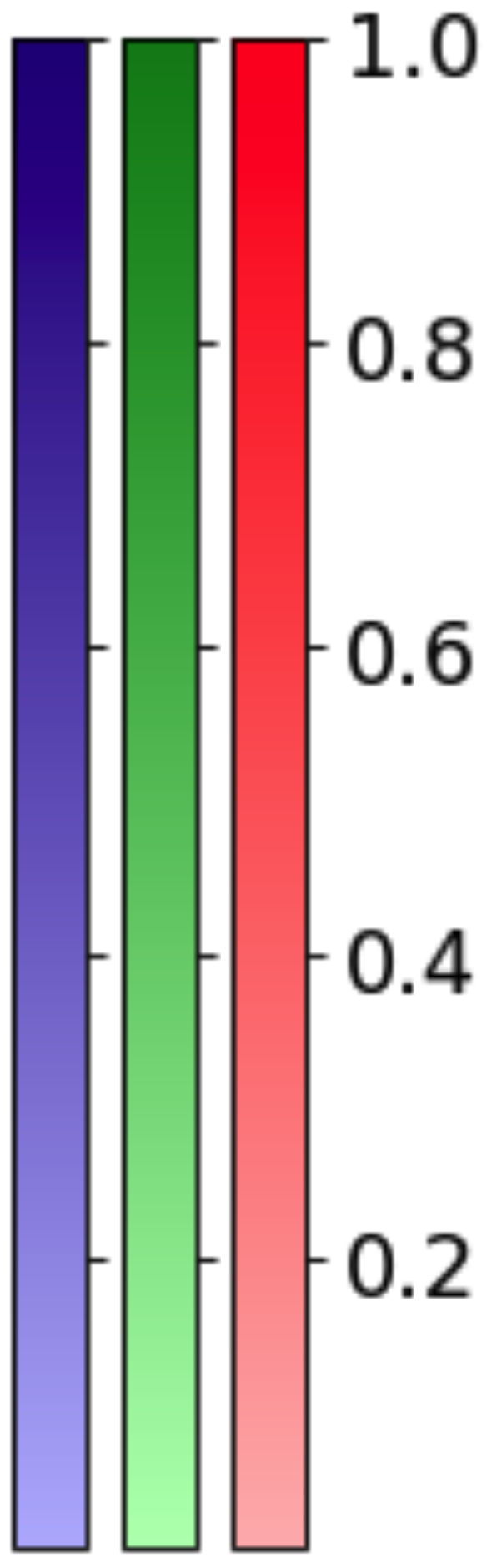}}
         \label{colorbars}
         \vspace{0.55cm}
     \end{subfigure}
     
     \caption{Illustration of the spatial distribution of activities (a),
     and the best sensor locations found in all runs of DGBO (b)
     and in all runs of BO (c). The first row shows T1 and the second row shows T2. 
     The intensity of the color shows the likelihood of installing a sensor at that location.}
     \label{final_placements}
\end{figure}

Second, Figure~\ref{final_placements} shows the best sensor locations found after 1,000 iterations 
in all runs of DGBO and BO (considering different $\epsilon$ values, random seeds, and target number of sensors) in T1 and T2.
It also shows the spatial distribution of activities using a heatmap overlaid 
on the floor plan of each building, with dark/light blue showing more/less activities in the space.
It can be readily seen that both methods place sensors in highly occupied areas, 
such as the kitchen and dining room.
Yet, DGBO's sensor placements are more promising. 
Specifically, in T1, BO places sensors in areas where no activities were performed
\eg the balcony, entryway, left side of the bedroom, and right side of the living room,
However, DGBO is less inclined to place sensors in these areas.
The same argument can be made for T2.

\begin{table*}[tb]
\centering
\footnotesize
\begin{tabular}{|p{0.4cm}|l|c|c|}
\hline
\multicolumn{2}{|c|}{\textbf{Testbed}} 
 & 
 \centering $100\times${\large $\frac{\textcolor[RGB]{0,128,0}{\bullet}{-}\textcolor[RGB]{200,0,0}{\bullet}}{\textcolor[RGB]{200,0,0}{\bullet}}$}  & 
 avg.  \\ \hline

\multirow{3}{*}{\textbf{T1}} & 
    \centering $\mathbf{\epsilon{=}0.25(m)}$     & 
    \centering $-17.9$\% &  
    \multirow{3}{*}{$-55.4\%$ } \\
    &
    \centering $\mathbf{\epsilon{=}0.5(m)}$ & 
    
    \centering $-61.8$\% & \\
    & 
    \centering $\mathbf{\epsilon{=}1(m)}$ & 
    
    \centering $-86.6\%$ & \\ \hline

\multirow{3}{*}{\textbf{T2}} & 
    \centering $\mathbf{\epsilon{=}0.25(m)}$ & 
    
    \centering $-41.0\%$ & \multirow{3}{*}{$-58.9\%$ } \\
    & 
    \centering $\mathbf{\epsilon{=}0.5(m)}$ & 
    
    \centering $-71.7\%$ & \\
    &
    \centering $\mathbf{\epsilon{=}1(m)}$ &
    
    \centering $-64.1\%$ & \\ \hline
        
\multicolumn{2}{|c|}{\textbf{Aruba}} &
    \centering $-39.6\%$ & $-39.6\%$ \\ \hline
\end{tabular}
\caption{The convergence analysis of DGBO compared to BO across our case studies. The avg. shows the average value of red/green dots of each testbed for all $\epsilon$ values.}
\label{conv}
\end{table*}

\begin{figure}[t!]
     \centering
     \begin{subfigure}{0.23\columnwidth}
         \centering
         \centerline{\includegraphics[width=\columnwidth]{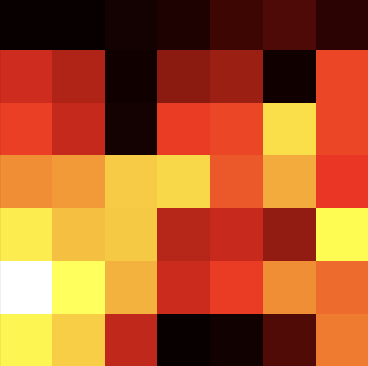}}
         \label{T1_EIG_50}
     \end{subfigure}
     \hspace{0.5em}
     \begin{subfigure}{0.28\columnwidth}
         \centering
         \centerline{\includegraphics[width=\columnwidth]{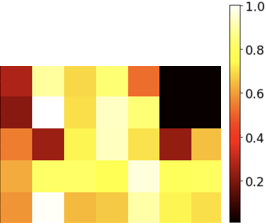}}
         \label{T2_EIG_50}
     \end{subfigure}
     \caption{Expected information gain: T1 (left), T2 (right).}
     \label{discuss}
\end{figure}

\section{Discussion}
\label{discussion}
We have found that DGBO and BO outperform the conventional methods for finding motion sensor placements, \ie  genetic and greedy algorithms, resulting in significantly higher activity recognition accuracy.
The DGBO learns the spatial distribution of activities during the search process and 
utilizes this distribution to consistently find high-quality sensor placements using noticeably fewer queries than BO.
We wish to emphasize that in the BO literature, $f(x)$ queries are typically costly~\cite{frazier2018tutorial}. 
In our problem, resolving these queries even through simulation is time-consuming, 
taking roughly 2 minutes on a computer with an Intel i7 4.00Ghz CPU and 16GB memory. 
Thus, coming up with sample efficient algorithms is key.

We argue that our spatial credit assignment strategy (Equation~\ref{info})
leads to an accurate estimation of the expected information gain of each region
after a small number of iterations.
To show this, we plot the expected information gain at iteration $n{=}50$
in Figure~\ref{discuss}.
Comparing this to Figure~\ref{final_placements} (b), we conclude that our estimation of expected information gain becomes accurate early in the optimization process.

We wish to clarify that our framework in Figure~\ref{fig_framework} is not designed 
for a particular choice of the activity classifier. 
To show this, we have repeated the process in T1 with $\epsilon{=}1$
using two different classifiers: 
gradient boosting and K-nearest neighbor (KNN) with $k{=}5$. 
In both cases, the superior performance of BO and DGBO remains statistically significant, and DGBO remains significantly more sample-efficient.

Further studies should be conducted using DGBO.
A future work direction is to apply transfer learning to the expected information gain 
and use this information in other environments.
Another future direction is to assess the effectiveness of DGBO in other emerging applications such as air pollution monitoring~\cite{hellan2022bayesian}, wildfire monitoring~\cite{gholami2021there}, 
and effective emergency response~\cite{ghosh2018dispatch}.
Finally, we have explored the optimal placement of one type of sensor
that is not deemed intrusive and is commonly used for activity detection in aging-in-place settings.
Our method can be extended to other sensor types, \eg mmWave radars.

\section{Conclusion}
\label{conclusion}
This paper casts optimal sensor placement in indoor environments as a black-box optimization problem
addressed using Bayesian Optimization (BO), a method that has not been applied to this problem before.
It then introduces Distribution-Guided Bayesian Optimization (DGBO) that incorporates 
the learned spatial distribution of activities into the acquisition function of Bayesian optimization.
Our approach entails using (a) a high-fidelity simulator for modeling the indoor environment, 
its occupants, and sensors; (b) an optimization algorithm 
to find a sensor placement that maximizes the detection accuracy of ADL.
We hypothesized that DGBO could explore the search space more effectively.
To test this hypothesis, we evaluated DGBO in two simulated suites of an assisted living facility and using one real-world dataset where subjects performed ADLs.
We compared the performance of DGBO, with BO and
two widely-used baselines, \ie genetic and greedy algorithms. 
Our result confirmed that DGBO  
finds high-quality sensor placements at a significantly lower cost.

\bibliographystyle{unsrt}  
\bibliography{MainFile}

\begin{thebibliography}{10}

\bibitem{cook2012casas}
Diane~J Cook, Aaron~S Crandall, Brian~L Thomas, and Narayanan~C Krishnan.
\newblock Casas: A smart home in a box.
\newblock {\em Computer}, 46(7):62--69, 2012.

\bibitem{pereira2018influence}
Pedro~F Pereira and Nuno~MM Ramos.
\newblock The influence of sensor placement in the study of occupant behavior
  in a residential building.
\newblock In {\em 2018 International Conference on Smart Energy Systems and
  Technologies (SEST)}, pages 1--6. IEEE, 2018.

\bibitem{yang2021adaptive}
Chen Yang.
\newblock An adaptive sensor placement algorithm for structural health
  monitoring based on multi-objective iterative optimization using weight
  factor updating.
\newblock {\em Mechanical Systems and Signal Processing}, 151:107363, 2021.

\bibitem{thomas2016genetic}
Brian~L Thomas, Aaron~S Crandall, and Diane~J Cook.
\newblock A genetic algorithm approach to motion sensor placement in smart
  environments.
\newblock {\em Journal of reliable intelligent environments}, 2(1):3--16, 2016.

\bibitem{wu2020sensor}
Hui Wu, Zhe Liu, Jin Hu, and Weifeng Yin.
\newblock Sensor placement optimization for critical-grid coverage problem of
  indoor positioning.
\newblock {\em International Journal of Distributed Sensor Networks},
  16(12):1550147720979922, 2020.

\bibitem{yu2020optimizing}
Xiaofan Yu, Kazim Ergun, Ludmila Cherkasova, and Tajana~{\v{S}}imuni{\'c}
  Rosing.
\newblock Optimizing sensor deployment and maintenance costs for large-scale
  environmental monitoring.
\newblock {\em IEEE Transactions on Computer-Aided Design of Integrated
  Circuits and Systems}, 39(11):3918--3930, 2020.

\bibitem{mori2005comparison}
Naoki Mori, Masayuki Takeda, and Keinosuke Matsumoto.
\newblock A comparison study between genetic algorithms and bayesian optimize
  algorithms by novel indices.
\newblock In {\em Proceedings of the 7th annual conference on genetic and
  evolutionary computation}, pages 1485--1492, 2005.

\bibitem{shahriari2015taking}
Bobak Shahriari, Kevin Swersky, Ziyu Wang, Ryan~P Adams, and Nando De~Freitas.
\newblock Taking the human out of the loop: A review of bayesian optimization.
\newblock {\em Proceedings of the IEEE}, 104(1):148--175, 2015.

\bibitem{deshwal2022bayesian}
Aryan Deshwal, Syrine Belakaria, Janardhan~Rao Doppa, and Dae~Hyun Kim.
\newblock Bayesian optimization over permutation spaces.
\newblock In {\em Proceedings of the AAAI Conference on Artificial
  Intelligence}, volume~36, pages 6515--6523, 2022.

\bibitem{deshwal2020optimizing}
Aryan Deshwal, Syrine Belakaria, Janardhan~Rao Doppa, and Alan Fern.
\newblock Optimizing discrete spaces via expensive evaluations: A learning to
  search framework.
\newblock In {\em Proceedings of the AAAI Conference on Artificial
  Intelligence}, volume~34, pages 3773--3780, 2020.

\bibitem{deshwal2023bayesian}
Aryan Deshwal, Sebastian Ament, Maximilian Balandat, Eytan Bakshy,
  Janardhan~Rao Doppa, and David Eriksson.
\newblock Bayesian optimization over high-dimensional combinatorial spaces via
  dictionary-based embeddings.
\newblock In {\em International Conference on Artificial Intelligence and
  Statistics}, pages 7021--7039. PMLR, 2023.

\bibitem{hellan2022bayesian}
Sigrid~Passano Hellan, Christopher~G Lucas, and Nigel~H Goddard.
\newblock Bayesian optimisation for active monitoring of air pollution.
\newblock In {\em Proceedings of the AAAI Conference on Artificial
  Intelligence}, volume~36, pages 11908--11916, 2022.

\bibitem{astudillo2021thinking}
Raul Astudillo and Peter~I Frazier.
\newblock Thinking inside the box: A tutorial on grey-box bayesian
  optimization.
\newblock In {\em 2021 Winter Simulation Conference (WSC)}, pages 1--15. IEEE,
  2021.

\bibitem{fanti2017smart}
Maria~Pia Fanti, Michele Roccotelli, Gregory Faraut, and Jean-Jacques Lesage.
\newblock Smart placement of motion sensors in a home environment.
\newblock In {\em 2017 IEEE International Conference on Systems, Man, and
  Cybernetics (SMC)}, pages 894--899. IEEE, 2017.

\bibitem{gungor2020respire}
Onat Gungor, Tajana~S Rosing, and Baris Aksanli.
\newblock Respire: Robust sensor placement optimization in probabilistic
  environments.
\newblock In {\em 2020 IEEE Sensors}, pages 1--4. IEEE, 2020.

\bibitem{barry2019computational}
Jordan Barry and Christopher Thron.
\newblock A computational physics-based algorithm for target coverage problems.
\newblock In {\em Advances in nature-inspired computing and applications},
  pages 269--290. Springer, 2019.

\bibitem{vlasenko2014smart}
Iuliia Vlasenko, Ioanis Nikolaidis, and Eleni Stroulia.
\newblock The smart-condo: Optimizing sensor placement for indoor localization.
\newblock {\em IEEE Transactions on Systems, Man, and Cybernetics: Systems},
  45(3):436--453, 2014.

\bibitem{weissteiner2023bayesian}
Jakob Weissteiner, Jakob Heiss, Julien Siems, and Sven Seuken.
\newblock Bayesian optimization-based combinatorial assignment.
\newblock In {\em Proceedings of the AAAI Conference on Artificial
  Intelligence}, volume~37, pages 5858--5866, 2023.

\bibitem{golestan2020towards}
Shadan Golestan, Ioanis Nikolaidis, and Eleni Stroulia.
\newblock Towards a simulation framework for smart indoor spaces.
\newblock {\em Sensors}, 20(24):7137, 2020.

\bibitem{briscoe2022adversarial}
Jarren Briscoe, Assefaw Gebremedhin, Lawrence~B Holder, and Diane~J Cook.
\newblock Adversarial creation of a smart home testbed for novelty detection.
\newblock In {\em AAAI Spring Symposium on Designing AI for Open Worlds}, 2022.

\bibitem{dahmen2019synsys}
Jessamyn Dahmen and Diane Cook.
\newblock Synsys: A synthetic data generation system for healthcare
  applications.
\newblock {\em Sensors}, 19(5):1181, 2019.

\bibitem{freedman2019unifying}
Richard~G Freedman and Shlomo Zilberstein.
\newblock A unifying perspective of plan, activity, and intent recognition.
\newblock In {\em Proceedings of the AAAI Workshops: Plan, Activity, Intent
  Recognition (Honolulu, HI)}, pages 1--8, 2019.

\bibitem{hutter2011sequential}
Frank Hutter, Holger~H Hoos, and Kevin Leyton-Brown.
\newblock Sequential model-based optimization for general algorithm
  configuration.
\newblock In {\em Learning and Intelligent Optimization: 5th International
  Conference, LION 5, Rome, Italy, January 17-21, 2011. Selected Papers 5},
  pages 507--523. Springer, 2011.

\bibitem{frazier2018tutorial}
Peter~I Frazier.
\newblock A tutorial on bayesian optimization.
\newblock {\em arXiv preprint arXiv:1807.02811}, 2018.

\bibitem{li2021openbox}
Yang Li, Yu~Shen, Wentao Zhang, Yuanwei Chen, Huaijun Jiang, Mingchao Liu,
  Jiawei Jiang, Jinyang Gao, Wentao Wu, Zhi Yang, et~al.
\newblock Openbox: A generalized black-box optimization service.
\newblock In {\em Proceedings of the 27th ACM SIGKDD Conference on Knowledge
  Discovery \& Data Mining}, pages 3209--3219, 2021.

\bibitem{sutton1984temporal}
Richard~Stuart Sutton.
\newblock {\em Temporal credit assignment in reinforcement learning}.
\newblock University of Massachusetts Amherst, 1984.

\bibitem{lifestyleoptions}
{Lifestyle Options}.
\newblock Lifestyle options.
\newblock \url{https://lifestyleoptions.ca/terra-losa/}, 2023.
\newblock Accessed: 2022-12-20.

\bibitem{gholami2021there}
Shahrzad Gholami, Narendran Kodandapani, Jane Wang, and Juan~Lavista Ferres.
\newblock Where there's smoke, there's fire: Wildfire risk predictive modeling
  via historical climate data.
\newblock In {\em Proceedings of the AAAI Conference on Artificial
  Intelligence}, volume~35, pages 15309--15315, 2021.

\bibitem{ghosh2018dispatch}
Supriyo Ghosh and Pradeep Varakantham.
\newblock Dispatch guided allocation optimization for effective emergency
  response.
\newblock In {\em Proceedings of the AAAI Conference on Artificial
  Intelligence}, volume~32, 2018.

\bibitem{krause2008efficient}
Andreas Krause, Jure Leskovec, Carlos Guestrin, Jeanne VanBriesen, and Christos
  Faloutsos.
\newblock Efficient sensor placement optimization for securing large water
  distribution networks.
\newblock {\em Journal of Water Resources Planning and Management},
  134(6):516--526, 2008.

\end{thebibliography}

\section{Appendix}
\subsection{Baselines}

\subsubsection{Genetic Algorithm (GA)}
We use the genetic algorithm used by Thomas~\textit{et al.}~\cite{thomas2016genetic}
to find an optimal motion sensor placement that supports accurate activity recognition. 
In this setting, every chromosome is a binary vector of size $L$, 
representing the locations of sensors in $x$.
The GA starts with a population of $10$ random sensor placements 
and feeds each one to the black-box function to obtain a fitness value. 
The fitness value of a chromosome is penalized by the number of sensors deployed 
(i.e. the number of $1$s in the chromosome) to prevent the algorithm from placing too many sensors.
The best $10\%$ of the chromosomes are kept, 
and crossover and mutation are performed to create the next generation.
First, the GA randomly chooses a pair of chromosomes 
from the best $20\%$ of the population in the crossover function. 
Then, it generates two distinct random indices, excluding the first and last indices, 
to slice each chromosome into three sections and swap their middle sections, 
generating two new offspring. 
Finally, it mutates each newly generated offspring by randomly choosing a $l_i{\in}\mathcal{L}$ and 
flipping its value with the probability of $0.005$. 

\begin{table*}[!t]
\scriptsize
\centering
\begin{tabular}{|p{0.7cm}|p{0.6cm}|p{1.2cm}|p{1.2cm}|p{1.2cm}|p{1.2cm}|p{1.2cm}|p{1.2cm}|p{1.2cm}|}

\hline
     & & \multicolumn{7}{c|}{\textbf{Avg. $F^1$ (${\pm}$ one standard deviation)}} \\ \cline{3-9}
     
    & &

    \multicolumn{3}{c|}{\textbf{T1}} &
    \multicolumn{3}{c|}{\textbf{T2}} &
    \multirow{2}{*}{\textbf{Aruba}} \\ \cline{3-8}

    {{\centering \rotatebox[origin=l]{0}{\textbf{Method}}}} & 
    {{\centering \rotatebox[origin=l]{0}{\textbf{Sens.~\#}}}} & 
    
    $\mathbf{ \epsilon{=}0.25(m)}$ &
    $\mathbf{\epsilon{=}0.5(m)}$ &
    $\mathbf{\epsilon{=}1.0(m)}$ &
    $\mathbf{\epsilon{=}0.25(m)}$ &
    $\mathbf{\epsilon{=}0.5(m)}$ &
    $\mathbf{\epsilon{=}1.0(m)}$ &
    \\

    \hline
     
    \textbf{GA} & N/A &
    $56.7{\pm}1.0$ (9)&
    $59.7{\pm}0.4$ (11) &
    $54.5{\pm}1.0$ (9) &
    $42.9{\pm}1.5$ (10) &
    $42.7{\pm}0.4$ (6) &
    $40.7{\pm}1.3$ (5) &
    $60.2{\pm}1.2$ (7)\\ 
    \hline

     \multirow{6}{*}{\textbf{Greedy}} & 
    5 & 
    ---  &
    --- & 
    $61.1{\pm}4.4$ &
     
    ---  &
    $59.0{\pm}2.0$ & 
    $56.5{\pm}3.8$ &
    $53.6{\pm}1.7$\\
    
     &
     7 &
     --- &
     --- &
     $61.2{\pm}3.2$ &
     
     --- &
     $59.8{\pm}2.7$ &
     $58.8{\pm}1.2$&
    $58.0{\pm}1.9$\\
     
     &
     9 &
     --- &
     --- &
     $57.3{\pm}3.1$ &
     
     --- &
     --- &
     $62.3{\pm}1.7$&
    $60.7{\pm}1.1$\\
     
     &
     11 &
     --- &
     --- &
     $67.1{\pm}3.4$ &
     
     --- &
     --- &
     $63.7{\pm}1.7$&
    $61.4{\pm}2.0$\\
     
     &
     13 &
     --- &
     --- &
     $69.7{\pm}1.1$ &

     ---  &
     --- &
     $65.1{\pm}0.8$&
    $63.0{\pm}1.4$\\
     
     &
     15 &
     --- &
     --- &
     $66.8{\pm}5.3$ &
     
     --- &
     --- &
     $66.8{\pm}1.3$&
    $64.0{\pm}1.3$\\ \hline
    
    \multirow{6}{*}{\textbf{BO}} & 
    5 & 
    $\mathbf{73.8{\pm}0.8}$  &
    $\mathbf{72.0{\pm}0.6}$ & 
    $68.4{\pm}4.4$ &

    $65.3{\pm}0.9$  &
    $\mathbf{64.7{\pm}0.3}$ & 
    $\mathbf{60.3{\pm}0.4}$ &
    $72.4{\pm}0.3$\\
    
     & 
    7 &
    $75.2{\pm}0.9$ &                
    $\mathbf{75.6{\pm}2.0}$ &                
    $72.8{\pm}1.2$ &

    $67.0{\pm}0.8$ &                
    $\mathbf{66.6{\pm}0.6}$ &                
    $63.1{\pm}0.4$ &
    $75.7{\pm}0.2$\\
    
     & 
    9 &
    $76.9{\pm}0.1$  &
    $72.4{\pm}0.8$ &
    $75.0{\pm}1.1$ &

    $\mathbf{68.9{\pm}0.5}$  &
    $67.3{\pm}0.7$ &
    $64.0{\pm}0.3$ &
    $74.8{\pm}1.4$\\
    
     & 
    11 & 
    $75.0{\pm}1.8$ & 
    $74.0{\pm}1.7$ & 
    $71.3{\pm}1.3$ &

    $72.3{\pm}1.6$  & 
    $\mathbf{67.7{\pm}0.3}$ & 
    $65.1{\pm}0.6$&
    $74.5{\pm}0.7$\\
    
     & 
    13 & 
    $73.3{\pm}0.7$ & 
    $\mathbf{74.9{\pm}0.9}$ & 
    $72.3{\pm}1.3$ &

    $68.4{\pm}0.6$ & 
    $\mathbf{67.5{\pm}1.0}$ & 
    $66.9{\pm}1.1$ &
    $74.4{\pm}0.5$\\
    
     & 
    15 & 
    $75.3{\pm}0.1$  & 
    $74.7{\pm}1.0$ & 
    $70.2{\pm}0.1$ &

    $68.4{\pm}1.4$  & 
    $67.0{\pm}0.6$ & 
    $67.1{\pm}0.8$ &
    $\mathbf{74.8{\pm}0.4}$\\
    
    \hline

    \multirow{6}{*}{\textbf{DGBO}} & 
    5 & 
    $67.5{\pm}2.4$  &
    $67.0{\pm}1.6$ & 
    $\mathbf{71.4{\pm}0.7}$ &
    
    $\mathbf{65.4{\pm}0.4}$  &
    $63.9{\pm}1.4$ & 
    $59.2{\pm}0.8$&
    $\mathbf{73.3{\pm}0.2}$\\
    
     & 
    7 &
    $\mathbf{75.8{\pm}1.9}$  &
    $73.5{\pm}0.6$ & 
    $\mathbf{76.8{\pm}0.8}$ &
    
    $\mathbf{69.8{\pm}1.3}$  &
    $66.1{\pm}1.0$ & 
    $\mathbf{64.6{\pm}1.8}$&
    $\mathbf{75.9{\pm}0.2}$\\
    
     & 
    9 &
    $\mathbf{77.6{\pm}1.1}$  &
    $\mathbf{74.3{\pm}0.5}$ & 
    $\mathbf{76.5{\pm}2.5}$ &
    
    $66.6{\pm}3.83$  &
    $\mathbf{68.5{\pm}1.3}$ & 
    $\mathbf{67.2{\pm}0.3}$&
    $\mathbf{76.3{\pm}0.2}$\\
    
     & 
    11 & 
    $\mathbf{76.8{\pm}0.6}$  &
    $\mathbf{75.4{\pm}0.5}$ & 
    $\mathbf{77.6{\pm}0.2}$ &
    
    $\mathbf{72.5{\pm}1.8}$  &
    $67.4{\pm}0.9$ & 
    $\mathbf{69.6{\pm}0.4}$&
    $\mathbf{76.1{\pm}0.1}$\\
    
     & 
    13 & 
    $\mathbf{76.3{\pm}1.1}$  &
    $74.8{\pm}2.6$ & 
    $\mathbf{73.5{\pm}1.1}$&
    
    $\mathbf{71.3{\pm}0.2}$  &
    $66.0{\pm}0.5$ & 
    $\mathbf{69.5{\pm}0.3}$&
    $\mathbf{74.7{\pm}0.2}$\\
    
     & 
    15 & 
    $\mathbf{75.42{\pm}0.9}$  &
    $\mathbf{77.51{\pm}0.1}$ & 
    $\mathbf{76.27{\pm}1.2}$ &
    
    $\mathbf{68.59{\pm}1.2}$  &
    $\mathbf{68.79{\pm}2.6}$ & 
    $\mathbf{68.7{\pm}1.2}$&
    $74.66{\pm}0.2$\\

\hline
\end{tabular}
\caption{The performance of GA, Greedy, BO and DGBO in terms of the macro-averaged $F^1$ in T1, T2 and Aruba for different sensor numbers.
A dash is used when a sensor configuration is not found using $1000$ queries. 
The best average performance for a given number of sensors is printed in bold.
We report in brackets the median number of sensors for GA.}
\label{app_resultsTable}
\end{table*}

\subsubsection{Greedy Algorithm}
We apply the natural greedy sensor placement choices described in~\cite{krause2008efficient}, 
wherein the algorithm makes the best decision in each iteration to optimize the entire problem. 
Specifically, it starts with a placement that contains no sensors,
and iteratively adds a sensor to this placement 
by finding the one-sensor placement that yields the highest $f(x)$ 
among the possible placements in $\mathcal{L}$.
The process continues until it places the given number of sensors 
or the maximum number of queries to $f(x)$ is reached. 
Notice that the greedy algorithm requires many queries to add a sensor to its current placement.
For example, for T1 with $\epsilon{=}0.25$ it requires 1115 function queries to place 5 sensors
with this breakdown: 225, 224, 223, 222, 221 queries for putting the 1st, 2nd, 3rd, 4th, and 5th sensor, respectively.
We note that the greedy algorithm may not find the global optimum 
as it is an iterative method and 
it overlooks that the observations of the black-box function are inherently noisy in each iteration.

\subsection{Additional Experiments}
\subsubsection{Non-triviality of the Sensor Placement Problem}
We give an example to explain why the problem of installing a fixed number of sensors 
to maximize the performance of an activity classification model is non-trivial.
Consider the $31$ motion sensors that were originally installed in the Aruba testbed.
The locations of these sensors were probably determined without solving a black-box optimization problem, 
so we examine the activity recognition accuracy using data generated by exactly these $31$ motion sensors.
Our experiment indicates that this manual sensor placement 
results in an average $F^1$ score of $67.8{\pm}0.5$ for $100$ runs.
This performance is markedly lower than that of the solutions found by DGBO and BO,
underscoring that trivial solutions are not optimal in most cases 
and a high quality solution can be identified through optimization only.

\subsubsection{Extended Performance Evaluation}
This section presents the complete evaluation results of DGBO, BO, and the two baselines.
Table~\ref{app_resultsTable} shows the performance of DGBO, BO, greedy, and GA
for different numbers of sensors. 
We also show the average performance of DGBO, BO, and greedy across 5 runs 
after each iteration with different sensor numbers. 
According to this table, after 1000 queries, 
DGBO performs better than BO, greedy, and GA in nearly three-fourths of the cases.
In the next section, we illustrate the better sample efficiency of DGBO 
by inspecting its performance after a different number of queries.

\begin{table*}[!t]
\centering
\scriptsize
\begin{tabular}{|c|c|c|c|c|c|c|c|}
\hline
&
\multicolumn{7}{c|}{\centering $100\times${\large $\frac{\textcolor[RGB]{0,128,0}{\bullet}{-}\textcolor[RGB]{200,0,0}{\bullet}}{\textcolor[RGB]{200,0,0}{\bullet}}$}} \\ \cline{2-8}
&
\multicolumn{3}{c|}{\textbf{T1}} &
\multicolumn{3}{c|}{\textbf{T2}} &
\multirow{2}{*}{\textbf{Aruba}} \\ \cline{2-7}

    \textbf{Sens \#}&
    $\mathbf{ \epsilon{=}0.25(m)}$ &
    $\mathbf{\epsilon{=}0.5(m)}$ &
    $\mathbf{\epsilon{=}1.0(m)}$ &
    $\mathbf{\epsilon{=}0.25(m)}$ &
    $\mathbf{\epsilon{=}0.5(m)}$ &
    $\mathbf{\epsilon{=}1.0(m)}$ & \\ \hline
    
    \textbf{5} &
    N/A &
    $10.3\%$ &
    $-78.1\%$ &
    $12.3\%$ &
    $-00.6\%$ &
    $1.3\%$ &
    $-52.2\%$ \\ \hline

    \textbf{7} &
    N/A &
    $15.4\%$ &
    $-81.0\%$ &
    $3.5\%$ &
    $-55.9\%$ &
    $20.1\%$ &
    $-10.9\%$ \\ \hline

    \textbf{9} &
    $-15.6\%$ &
    $-39.6\%$ &
    $-49.1\%$ &
    $26.0\%$ &
    $-47.5\%$ &
    $-51.7\%$ &
    $-54.6\%$ \\ \hline

    \textbf{11} &
    $-6.2\%$ &
    $-26.3\%$ &
    $-91.7\%$ &
    $-41.0\%$ &
    $-54.7\%$ &
    $-20.1\%$ &
    $-90.4\%$ \\ \hline

    \textbf{13} &
    $-95.3\%$ &
    $-84.2\%$ &
    $-95.3\%$ &
    $-74.4\%$ &
    $-43.6\%$ &
    $-72.7\%$ &
    $-84.3\%$ \\ \hline

    \textbf{15} &
    $-60.8\%$ &
    $-79.7\%$ &
    $-93.3\%$ &
    $-95.4\%$ &
    $-89.7\%$ &
    $-77.5\%$ &
    $-49.4\%$ \\ \hline

    \textbf{avg.} &
    $-29.3\%$ &
    $-42.9\%$ &
    $-81.4\%$ &
    $-28.1\%$ &
    $-48.6\%$ &
    $-40.5\%$ &
    $-49.6\%$ \\ \hline
\end{tabular}
\caption{The convergence analysis of DGBO compared to BO across our case studies.}
\label{conv_complete}
\end{table*}

\begin{figure*}[!bt]
    \centering
     \includegraphics[width=1\textwidth]{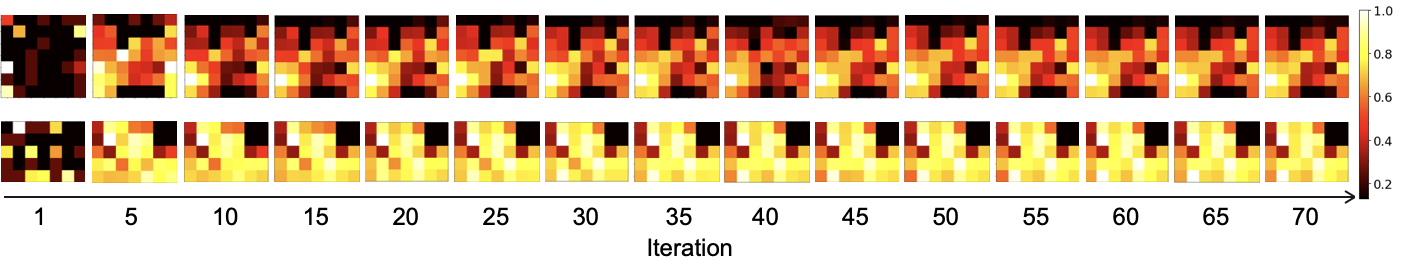}
     \caption{Expected information gain during the initial iterations of the optimization process for T1 (top) and T2 (bottom).}
     \label{app_eig2}
\end{figure*}

\subsubsection{A Closer Look at DGBO's Sample Efficiency}
In Table~\ref{conv_complete}, we provide the same result that was presented in the paper for different numbers of sensors. 
Concretely, for a given number of sensors, 
we find the first iteration at which DGBO and BO reach the $95\%$ confidence interval of the best performance of DGBO after 1,000 iterations. 
It can be readily seen that DGBO excels in terms of sample efficiency when 9 or more sensors are installed. 
We believe that this occurs because in Equation 8, the information profiles get updated using data from fewer activation regions when examining sensor placements with a smaller number of sensors; 
thus, the expected information gain converges slowly.

\subsubsection{Expected Information Gain Convergence}
Finally, verify the efficacy of the method proposed to calculate the expected information gain.
To this end, we take a look at how the expected information gain converges 
to an accurate estimation of the spatial distribution of the activities. 
Figure~\ref{app_eig2} visualizes the expected information gain when DGBO is run for T1 and T2 with $\epsilon{=}1$ and 11 sensors. It can be seen that after about $50$ iterations, 
the relative changes to the expected information gain of each activation region $R_i$ become negligible.

\end{document}